\documentclass[sigconf]{acmart}
\AtBeginDocument{%
  }

\usepackage{amsmath}
\usepackage{multirow} 
\usepackage{tabularx}
\usepackage{marvosym}
\usepackage{colortbl}
\usepackage{pifont}
\usepackage{makecell}
\usepackage{booktabs}
\usepackage{balance}
\usepackage{graphicx}
\usepackage{adjustbox}
\usepackage[table]{xcolor}
\newcommand{\cmark}{\checkmark}
\newcommand{\xmark}{\texttimes}

\definecolor{statusgray}{gray}{0.88}
\definecolor{unitgray}{gray}{0.45}

\newcommand{\subgray}[1]{{\color{unitgray}\scriptsize #1}}

\newcommand{\headtwo}[2]{%
  \makecell[c]{\textbf{#1}\\[-0.2ex]\subgray{#2}}%
}

\newcommand{\grouptwo}[2]{%
  \makecell[c]{\textbf{#1}\\[-0.35ex]\subgray{#2}}%
}

\newcolumntype{C}[1]{>{\centering\arraybackslash}m{#1}}

\newcommand{\best}[1]{\textbf{#1}}
\newcommand{\second}[1]{\underline{#1}}
\newcommand{\NAcell}{\cellcolor{statusgray}\textcolor{black}{\textbf{N/A}}}

\setcopyright{acmlicensed}
\copyrightyear{2026}
\acmYear{2026}
\acmDOI{XXXXXXX.XXXXXXX}
\acmConference[Conference acronym 'XX]{Make sure to enter the correct
  conference title from your rights confirmation email}{June 03--05,
  2018}{Woodstock, NY}
\acmISBN{978-1-4503-XXXX-X/2018/06}

\acmSubmissionID{1736}

\begin{document}

\title{DepthART: Scaling Foundation Monocular Depth to Tiny Models}

\author{Feng Xue}
\orcid{0000-0002-4101-3401}
\affiliation{%
  \institution{University of Trento}
  \city{Trento}
  \country{Italy}}
\email{feng.xue@unitn.it}

\author{Wu Chen}
\affiliation{%
  \institution{Beijing University of Posts and Telecommunications}
  \city{Beijing}
  \country{China}}
\email{chenw@bupt.edu.cn}

\author{Mingshuai Zhao}
\orcid{0009-0001-7846-6391}

\author{Guofeng Zhong}

\affiliation{%
  \institution{Beijing University of Posts and Telecommunications}
  \city{Beijing}
  \country{China}}

\author{Anlong Ming}
\authornote{Corresponding author.}
\orcid{0000-0003-2952-7757}
\affiliation{%
  \institution{Beijing University of Posts and Telecommunications}
  \city{Beijing}
  \country{China}}
\email{mal@bupt.edu.cn}

\author{Haozhe Wang}
\affiliation{%
  \institution{The Hong Kong University of Science and Technology}
  \city{Hong Kong}
  \country{China}}
\email{jasper.whz@outlook.com}

\author{Dianqiao Lei}
\affiliation{%
  \institution{Tsinghua University}
  \city{Beijing}
  \country{China}}

\author{Zhaowen Lin}

\author{Haiyang Zhang}

\affiliation{%
  \institution{Beijing University of Posts and Telecommunications}
  \city{Beijing}
  \country{China}}

\author{Nicu Sebe}
\orcid{0000-0002-6597-7248}
\affiliation{%
  \institution{University of Trento}
  \city{Trento}
  \country{Italy}}
\email{niculae.sebe@unitn.it}

\renewcommand{\shortauthors}{Xue et al.}


\begin{abstract}
Recent geometric foundation models (\textit{e.g.}, \textit{Metric3D}, \textit{Depth Anything} and \textit{UniDepth}) have substantially improved monocular depth estimation (MDE) in both cross-scene generalization and metric-scale prediction,
yet these gains have not translated to tiny models.
We bridge this gap with \textbf{\textit{DepthART}} (\textbf{Depth} \textbf{A}nything \textbf{R}ethought for \textbf{T}iny Models),
which is a compact MDE model for on-device deployment across diverse scenes.
We first identify two capacity-driven bottlenecks in tiny models:
(i) overfitting to dataset-specific distribution bias and (ii) unstable metric adaptation under camera shift, where full fine-tuning easily damages transferable geometry.
Accordingly, DepthART combines two simple but effective strategies:
a bias-resistant data sampling scheme to reduce distribution bias under the same training budget,
and a camera-conditioned fine-tuning protocol that freezes the distilled encoder and adjusts metric scale conditioned on intrinsics while better preserving cross-dataset generalization.
Across datasets, DepthART consistently surpasses previous tiny baselines in both zero-shot generalization and metric accuracy (\textit{e.g.}, zero-shot $\delta_1\!\!=\!\!0.964$ for DepthART-S on NYUD v2),
and in some cases approaches heavy models.
We further provide a scalable model family,
with DepthART-S reaching \textbf{\textit{347/245}} FPS (strict FP32) on an RTX A6000 at $224^2$/$448^2$, \textbf{\textit{102}} FPS (TF32) on a
Orin NX 8GB, and over \textbf{\textit{15}} FPS (FP32) on a Jetson Nano 4GB.
Code is released on \hyperlink{https://github.com/xuefeng-cvr/DepthAR T}{\textit{GitHub}}. Project page: \hyperlink{https://xuefeng-cvr.github.io/DepthART}{\textit{xuefeng-cvr.github.io/DepthART}}.
\end{abstract}

\begin{teaserfigure}
  \includegraphics[width=\textwidth]{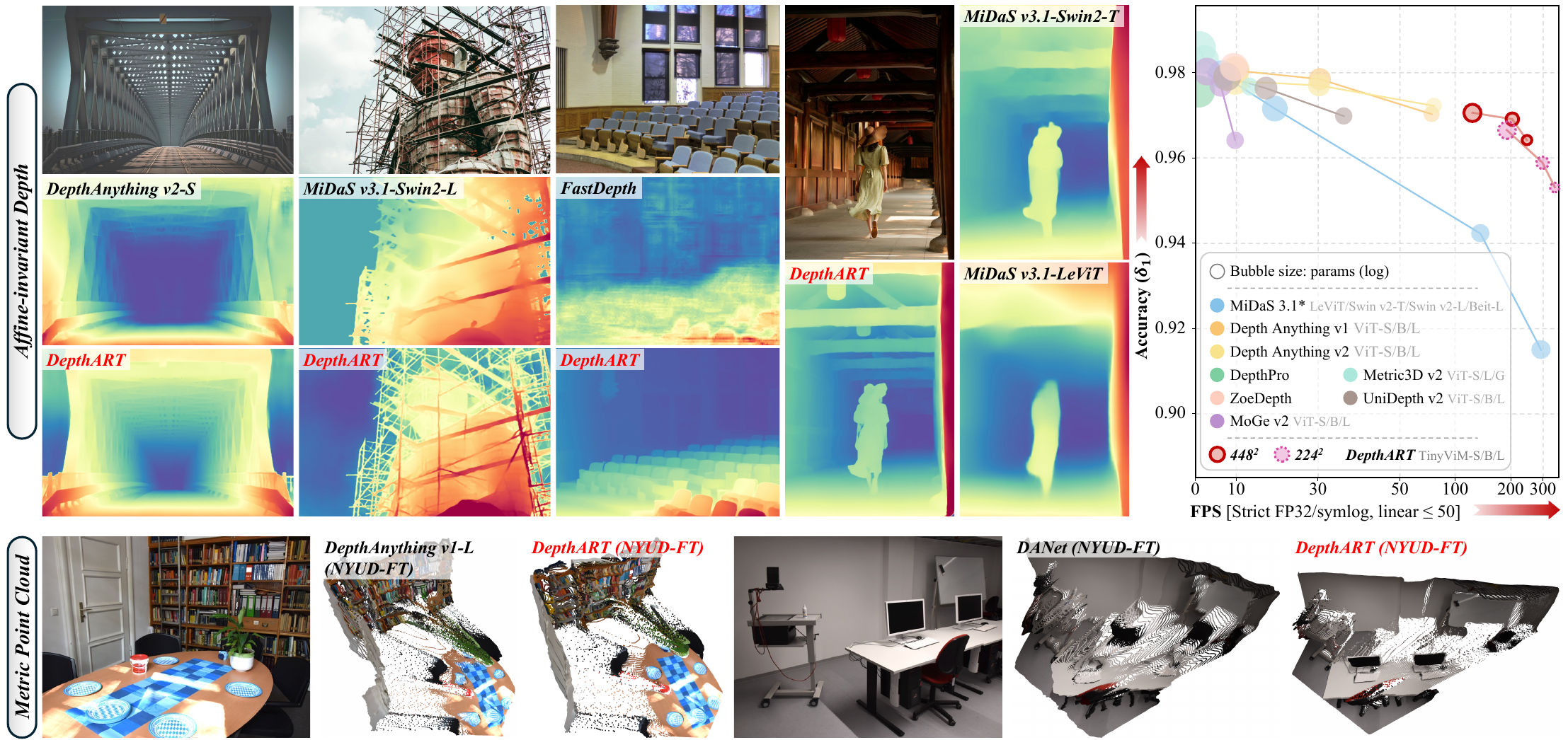}
\caption{
DepthART brings Depth Anything-style capability to tiny models,
achieving strong zero-shot generalization, reliable relative depth,
stable metric scale, and high efficiency.
DANet and FastDepth are lightweight models trained on NYUD v2.
Depth Anything v2-S uses a ViT-S encoder for relative depth,
while Depth Anything v1-L is fine-tuned on NYUD v2.
Affine-invariant depth is evaluated by per-image least-squares scale-and-shift fitting over valid pixels,
whereas metric 3D point clouds are visualized \textit{without any re-scaling} after fine-tuning.
The plot compares standard-protocol NYUD v2 $\delta_1$ with strict-FP32 model-only FPS.
DepthART is profiled at $224^2$ and $448^2$;
other methods use their standard runtime resolutions.
}
  \label{fig:teaser}
\end{teaserfigure}


\settopmatter{authorsperrow=4}
\maketitle

\section{Introduction}
Monocular depth estimation (MDE) aims to recover 3D scene geometry from a single RGB image and serves as a core geometric primitive for downstream multimedia systems,
such as image/video synthesis \cite{zhang2023controlnet,zavadski2024controlnetxs,kang2025look,zhang2018depth},
world models \cite{chai2023persistentnature,ren2025gen3c},
mobile robots \cite{merrill2024fast,yokoyama2024vlfm},
and augmented reality \cite{lee2025imaginatear}.
As these applications move from controlled settings to real-world deployment,
they increasingly require not only visually plausible depth,
but also reliable generalization across diverse scenes and cameras.

To meet this demand,
MDE has rapidly shifted toward foundation-model paradigms \cite{liu2024sm4depth,depth_anything_v2,hu2024metric3dv2,piccinelli2025unidepthv2,zeng2025depthdark}.
Early efforts improved generalization by training on broader mixtures (MiDaS \cite{ranftl2022midas}) and adopting stronger transformer backbones (DPT \cite{ranftl2021dpt}).
More recent work narrows the gap between relative-depth robustness and metric-scale prediction by scaling data and capacity and by introducing explicit camera modeling (Metric3D v1/v2 and UniDepth v1/v2 \cite{yin2023metric3d,hu2024metric3dv2,piccinelli2024unidepth,piccinelli2025unidepthv2}) or large-scale pseudo labels (Depth Anything v1/v2 \cite{depthanything,depth_anything_v2}).
Some methods further expand the camera model beyond pinhole (UniK3D \cite{piccinelli2025unik3d}) or predict richer 3D representations (MoGe v1/v2 \cite{wang2025moge,wang2025moge2}).
In parallel, generative priors offer a complementary route to robust dense geometry:
diffusion-based approaches (\textit{e.g.}, Marigold \cite{ke2023marigold} and GeoWizard \cite{fu2024geowizard}) often yield sharper structures and stronger robustness under distribution shift,
and recent works accelerate inference via one-step transfer \cite{xu2024genpercept,he2025lotus2} or flow matching \cite{gui2025depthfm}.
Together, these advances substantially raise the ability ceiling of MDE.

However, these foundation capabilities have not become standard for tiny models that must run in real time on resource-limited devices.
Rather than proposing a new camera-aware foundation depth architecture,
we study a different regime:
\textbf{\textit{how to preserve foundation-style transfer in tiny deployment-oriented backbones}}.
We find that naively shrinking foundation MDE fails for two regime-specific reasons:
\textbf{(i)} \textbf{\textit{Overfitting to dataset-specific bias}}:
When trained on mixed data sources, tiny models are more likely to overfit dataset-specific bias (\textit{e.g.}, frequent patterns) rather than learn a depth prior that transfers across scenes.
\textbf{(ii)} \textbf{\textit{Unstable metric adaptation under camera shift}}:
Tiny models often rely on implicitly encoded camera cues. As a result, full fine-tuning on a single metric dataset may improve in-domain accuracy, but it transfers poorly across cameras and more easily damages previously learned scene geometry.

To address this gap,
we propose DepthART,
a lightweight MDE model designed for real-time, on-device inference across diverse scenes.
DepthART is built on two simple strategies aligned with the failure modes above.
We first introduce a bias-resistant data sampling (BRDS) scheme.
BRDS explicitly re-balances the long-tailed distribution of photographic and camera-induced cues that tend to dominate depth learning in ultra-light models,
consistently improving cross-dataset zero-shot transfer under the same data budget.
In Stage 1, we distill from Depth Anything v2 to transfer foundation-style relative depth priors to a tiny student.
Then we design a camera-conditioned fine-tuning (CamFT) protocol.
CamFT freezes the distilled encoder and inserts lightweight intrinsics-conditioned adapters plus a multi-query scale head to recover metric scale on the target domain,
improving scale stability while substantially better preserving zero-shot geometry than full fine-tuning.
Across extensive evaluations,
DepthART substantially improves both scene generalization and metric-scale transfer over prior lightweight approaches,
and in some settings approaches the performance of much larger models,
as shown in the comparison of Fig. \ref{fig:teaser}.
We also provide multiple model sizes to match diverse deployment constraints,
highlighting the practicality under real-time, resource-limited conditions.
Our contributions are summarized as follows:
\begin{itemize}
\item  We identify a practically important but underexplored regime: transferring foundation-style monocular depth generalization to tiny deployment-oriented models.
\item We show that tiny MDE models are particularly vulnerable to dominant training distributions during distillation and to forgetting during metric fine-tuning, and address these issues with BRDS and encoder-frozen CamFT, respectively.
\item With a low-latency TinyViM+DPT backbone (6M/11M/32M), DepthART substantially advances both cross-dataset relative-depth generalization and metric fine-tuning, offering a practical recipe for resource-limited deployment.
\end{itemize}

\begin{figure*}
    \centering
    \includegraphics[width=\linewidth]{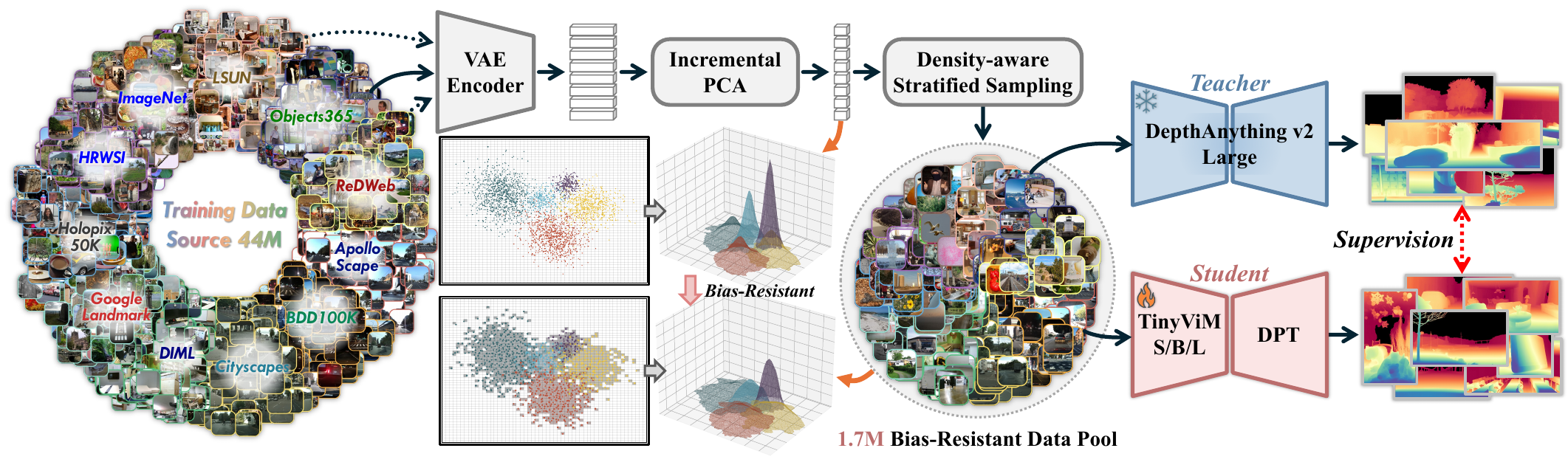}
    \caption{Bias-resistant data sampling (BRDS) and distillation pipeline.
    A 44M multi-source corpus is encoded (VAE) and reduced (Incremental PCA) to enable density-aware sample selection, producing a 1.7M bias-resistant training pool. The scatter plots and density surfaces illustrate the effect of rebalancing in feature space and are simulated for visualization. DepthART (TinyViM-S/B/L + DPT) is then distilled from Depth Anything v2 Large using teacher depth supervision.}
    \label{fig:pretrain}
\end{figure*}

\section{Related Work}
\noindent
\textbf{Monocular Depth Estimation} (MDE) \cite{bsnet,eigen2015predicting,zheng2018net, su2019monocular, wu2023v2depth, bhat2023zoedepth} was mainly studied in the single-domain supervised setting,
focusing on architecture and loss design to recover scene structure from one dataset.
Representative examples include DORN \cite{dorn}, VNL\cite{vnl}, and AdaBins \cite{adabin}, etc.
These approaches achieved strong in-domain accuracy,
but their generalization was often limited by the training data and camera settings.
Recent progress is driven by foundation training.
Methods such as MiDaS \cite{ranftl2022midas} and DPT \cite{ranftl2021dpt} improve zero-shot relative depth through large-scale, mixed-data pretraining. Metric3D\cite{hu2024metric3dv2} and UniDepth\cite{piccinelli2025unidepthv2} further target metric depth and camera variation.
UniK3D\cite{piccinelli2025unik3d} and MoGe \cite{wang2025moge2} move toward more explicit 3D reasoning.
Diffusion-based approaches (\textit{e.g.}, Lotus\cite{he2025lotus2}, Marigold\cite{ke2023marigold}, GeoWizard\cite{fu2024geowizard}) produce high-quality, detail-preserving depth (see introduction for more details).
However, lightweight MDE methods, including FastDepth\cite{wofk2019fastdepth}, DANet\cite{sheng2022danet}, GuideDepth\cite{guidedepth}, and LMDepth\cite{lmdepth}, are still trained and evaluated within a single domain, but with more efficient structure design.
Therefore, they generally lag behind foundation models.
DepthART is motivated by this gap:
rather than proposing a new camera-aware foundation depth formulation,
it aims to transfer foundation-level generalization and metric-scale robustness to compact models for on-device deployment.

\noindent
\textbf{Coreset Selection} typically compresses training data by selecting a small \emph{representative} subset that approximates full-data training, \textit{e.g.}, coverage-based \emph{k}-center selection in feature space~\cite{Sener2018Coreset} or gradient/submodular matching methods such as CRAIG and Grad-Match~\cite{Mirzasoleiman2020CRAIG,Killamsetty2021GradMatch}.
In monocular depth,
distribution bias has been explicitly identified as a key obstacle to cross-domain generalization,
\textit{e.g.}, DME~\cite{Yu2024DME} analyzes and mitigates depth-related bias.
Inspired by both coreset selection and bias-focused depth studies, our sampling resembles coverage-based selection but targets debiasing rather than dataset compression,
by suppressing over-dominant modes and promoting broad feature-space coverage for higher generalization.

\section{Method}
\subsection{Overview of DepthART}
DepthART targets tiny deployment-oriented MDE,
with an emphasis on preserving cross-scene transfer and enabling stable metric calibration under strict model budgets.
For efficiency,
DepthART adopts a compact Mamba-style encoder, \textit{i.e.}, TinyViM \cite{ma2025tinyvim},
paired with a widely used DPT decoder \cite{ranftl2021dpt},
forming a lightweight trunk.

The training proceeds in two stages.
In the first stage, we distill DepthART from a large foundation MDE model, \textit{e.g.}, Depth Anything v2,
to inherit robust cross-scene affine-invariant depth generalization.
As tiny models are highly sensitive to dataset bias,
we employ a bias-resistant data sampling scheme (\textbf{\textit{see Sec. \ref{sec:BRDSS}}}) to construct a bias-resistant subset from a large multi-source corpus for training.
In the second stage,
we freeze the distilled encoder and perform camera-conditioned fine-tuning (\textbf{\textit{see Sec. \ref{sec:camft}}}) on a single metric dataset, \textit{i.e.}, NYUD v2 or KITTI.
To mitigate catastrophic forgetting,
we insert a simple camera adapter to adjust the depth feature conditioned on camera intrinsics,
and a multi-query scale head to regress the entire scale of images.

\subsection{Bias-Resistant Relative Depth Pre-training}
\label{sec:BRDSS}

\vspace{5pt}\noindent\textbf{Dataset Distribution Bias Analysis.}
In practice,
manually collected datasets follow implicit priors shaped by \textit{geometry}, \textit{viewpoint}, and \textit{photometric} conditions.
For example,
\textbf{\textit{ImageNet}} \cite{imagenet} is largely object-centric,
\textbf{\textit{LSUN}} \cite{yu15lsun} often captures indoor scenes from similar camera viewpoints,
and \textbf{\textit{Google Landmark}} \cite{weyand2020googlelandmark} primarily focuses on landmarks, buildings, and natural landscapes,
exhibiting a typical ``tourist photography'' bias and a very long-tailed category distribution.
\textbf{\textit{BDD100K}} \cite{yu2020bdd100k}, in contrast, exhibits strong ego-pose priors from vehicle-mounted cameras.
However, comprehensive data coverage for MDE is more critical than tightly fitting frequent patterns of a single dataset.
Therefore,
it is essential to increase feature-space coverage and reduce the over-representation of high-density regions so that long-tail cases receive sufficient training exposure,
especially for capacity-limited models.

\vspace{5pt}\noindent\textbf{Bias-Resistant Data Sampling Strategy.}
Given an RGB image set $\mathcal{D}\!=\!\{x_i\}_{i=1}^{N}$ from a dataset,
our bias-resistant sampling applies density-aware stratified sub-sampling in an image-level feature space where similar images form clusters,
down-weighting dominant modes to retain diversity (Fig.~\ref{fig:pretrain}).

Specifically, for each image $x_i$,
we extract a continuous representation $z_i\!\!=\!\!E(x_i)$ using a well-trained variational autoencoder (VAE) $E(\cdot)$ from Stable Diffusion v1.5 \cite{stablediffusion}.
Unlike object-centric embeddings, \textit{e.g.}, DINO,
VAE latents tend to emphasize holistic visual factors (\textit{e.g.}, layout, texture, illumination, and viewpoint) that are more aligned with bias in training depth.
Secondly,
we project $z_i$ to a $J$-dimensional vector $\tilde{z}_i \in\mathbb{R}^{J}$ via Principal Component Analysis (PCA),
denoted as $\tilde{z}_i =\mathtt{PCA}(z_i)$,
which yields a compact and smoothly structured embedding space suitable for large-scale sampling.
Thirdly, to assign each sample to a grid cell, we perform per-dimension binning on $\tilde{z}_i$.
For each dimension $j \in \{1,\ldots,J\}$,
we partition its value range into $B$ bins with uniformly spaced boundaries $\{b^{(j)}_{k}\}_{k=0}^{B}$,
where $b^{(j)}_{0} \!=\! \min_i \tilde{z}^{(j)}_i$ and $b^{(j)}_{B}\!=\!\max_i \tilde{z}^{(j)}_i$.
We then compute $J$-D cell identifier, denoted as $c_i$, for an image $x_i$:
\begin{equation}
c_i = \left(c^{(1)}_i,...,c^{(J)}_i\right),\,
c_i^{(j)} = \max\left\{k \in [0,B-1] \big|\ b_k^{(j)} \le\tilde z_i^{(j)} \right\},
\end{equation}
Images sharing the same $c_i$ fall into the same grid cell and are treated as one stratum for subsequent density-aware sub-sampling.
Subsequently,
we perform a stratified sampling per cell.
Let $\mathcal{D}_\mathbf{c}\!=\!\{x_i \!\!\mid\!\! c_i\!=\!\mathbf{c}\}$ denote the images that fall into cell $\mathbf{c}$,
the sub-sampled set, denoted as $\mathcal{S}_\mathbf{c}\!\subset\! \mathcal{D}_\mathbf{c}$, is formulated as:
\begin{equation}
\mathcal{S}_\mathbf{c} \sim \mathrm{U}(\mathcal{D}_\mathbf{c}, m_\mathbf{c}), \quad m_\mathbf{c} = \max\left(1,\left\lceil \,|\mathcal{D}_\mathbf{c}|\,/\gamma\,\right\rceil\right)\!, 
\end{equation}
where $\gamma\!>\!1$ is the reduction factor,
$\mathrm{U}(\mathcal{D}_\mathbf{c}, m_\mathbf{c})$ denotes uniformly sub-sampling $m_\mathbf{c}$ images from $\mathcal{D}_\mathbf{c}$.
This rule enforces two properties:
(i) \textbf{Coverage}: even sparse, underrepresented regions remain visible in $\mathcal{S}_\mathbf{c}$ via $\max(1,\cdot)$ and (ii) \textbf{Bias reduction}: for dense, dominant regions, only a fraction of samples are retained, suppressing their influence.
Finally, we aggregate all selected samples into a bias-resistant subset:
\begin{equation}
\mathcal{D}_{\mathrm{BR}} = \bigcup\nolimits_{\mathbf{c}}\mathcal{S}_\mathbf{c}.
\end{equation}
In this way,
$\mathcal{D}_{\mathrm{BR}}$ preserves broad diversity across scenes and camera poses while suppressing the dominance of overrepresented modes in the original datasets.

\vspace{5pt}\noindent\textbf{Distillation from Foundation MDE Models.}
We distill DepthART on the offline sub-sampled set $\mathcal{D}_{\mathrm{BR}}$ using \emph{Depth Anything v2 Large} as the teacher.
For each image in $\mathcal{D}_{\mathrm{BR}}$ (across all training datasets), the teacher produces the pseudo depth and the student predicts the corresponding depth.
We follow a standard MDE distillation \cite{ranftl2022midas}:
\begin{equation}
\mathcal{L}_{\mathrm{Distill}}
= \mathcal{L}_{\mathrm{MSE}} + \alpha\,\mathcal{L}_{\mathrm{Edge}},
\qquad \alpha = 0.5,
\end{equation}
where $\mathcal{L}_{\mathrm{MSE}}$ measures the error between the teacher pseudo-depth and the student prediction after \emph{per-image scale/shift alignment}, and
$\mathcal{L}_{\mathrm{Edge}}$ encourages consistent depth discontinuities via an $L_{1}$ penalty on spatial gradients.
For higher robustness,
$\mathcal{L}_{\mathrm{MSE}}$ ignores the top-10\% highest-error pixels in each image,
preventing a few noisy pixels from dominating the distillation signal.

\begin{figure}[t]
\centering
\includegraphics[width=\linewidth]{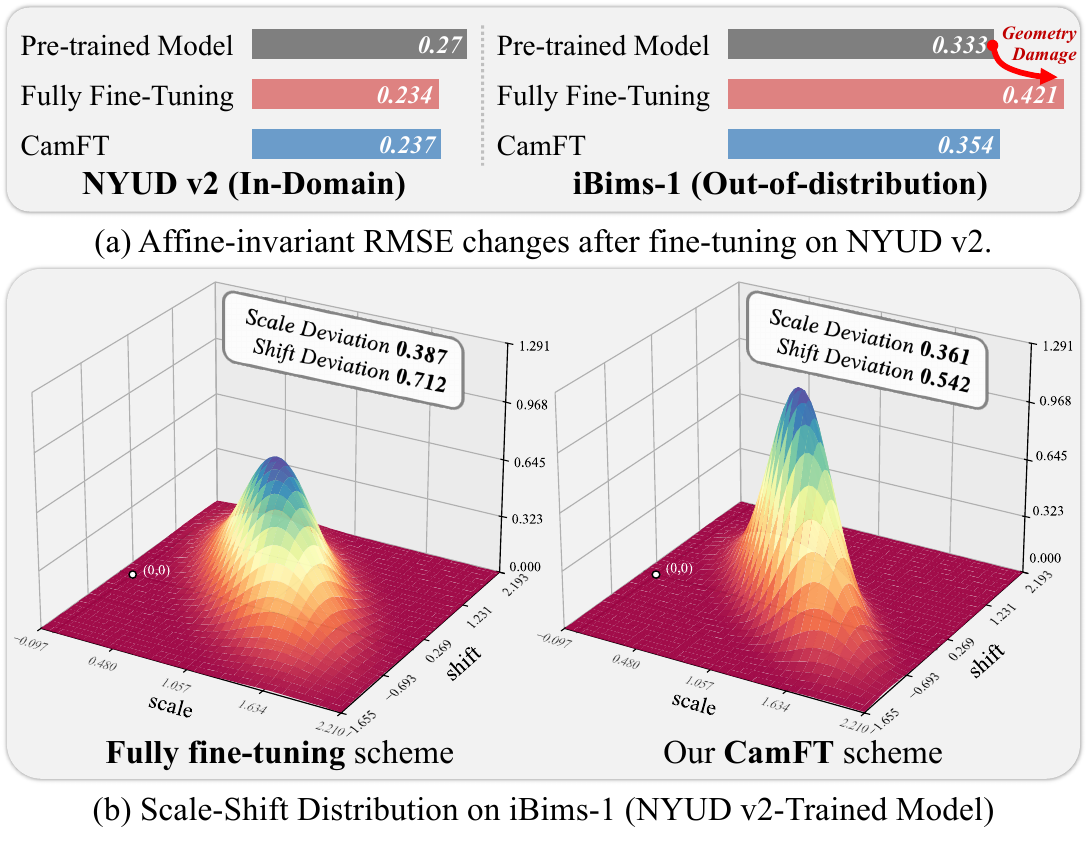}
\vspace{-1em}
\caption{
Geometry and scale stability after fine-tuning on NYUD v2.
(a) Full fine-tuning hurts zero-shot affine-invariant RMSE more than CamFT, indicating greater geometry damage.
(b) On iBims-1, CamFT produces a tighter distribution of per-image scale--shift alignment parameters, indicating less variable per-image affine corrections than full fine-tuning.
}
\label{fig:scale_shift_distribution}
\vspace{-4pt}
\end{figure}

\subsection{Camera-conditioned Metric Fine-tuning}
\label{sec:camft}

\vspace{5pt}\noindent\textbf{Forgetting Effects in Tiny Models Fine-Tuning.}
Due to limited capacity,
fully fine-tuning tiny models on a single metric dataset might easily overwrite the transferable depth cues learned via distillation, causing \emph{catastrophic forgetting}.

We employ affine-invariant metric to measure the degradation of depth structure after fine-tuning.
As shown in Fig. \ref{fig:scale_shift_distribution}(a),
full fine-tuning on NYUD v2 improves in-domain accuracy (Affine-invariant RMSE: $0.27\!\rightarrow\!0.234$) but substantially degrades zero-shot performance on iBims-1 ($0.333\!\rightarrow\!0.421$),
indicating that the depth structure of zero-shot scenes is damaged when all parameters are updated.
Beyond structural degradation, fine-tuning also perturbs metric-scale.
Fig. \ref{fig:scale_shift_distribution}(b) shows increased variability in the per-image scale and shift alignment parameters on iBims-1 after full fine-tuning ($\mathtt{std}: 0.387/0.712$), reflecting unstable scale transfer.

\begin{figure*}[t]
    \centering
    \includegraphics[width=\linewidth]{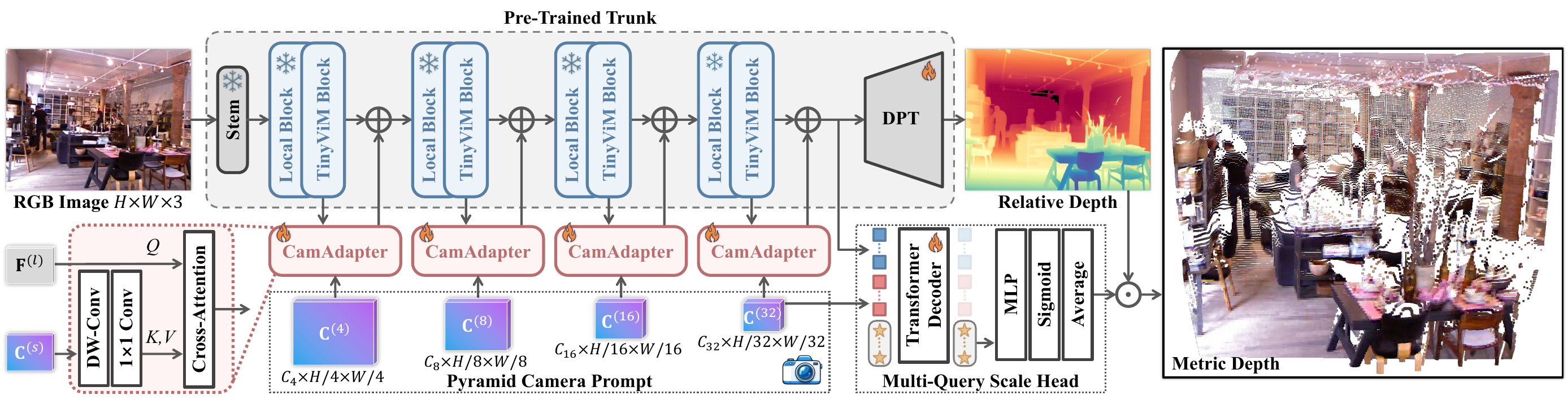}
    \caption{Camera-conditioned fine-tuning.
    We freeze the distilled encoder and inject lightweight camera adapters conditioned on pyramid camera prompts.
    A multi-query scale head predicts a scale factor,
    transferring the relative depth to metric depth.}
    \label{fig:finetune}
\end{figure*}

\vspace{5pt}\noindent\textbf{Camera-conditioned Fine-Tuning.}
To adapt metric scale without overwriting the pre-trained depth prior,
we perform a calibration-style fine-tuning that keeps the distilled encoder frozen (Fig.~\ref{fig:finetune}).
Instead of fully fine-tuning the model,
which easily overfits the target domain and damages learned geometry in tiny models,
we add two trainable modules:
(i) \textit{camera adapters} that modulate encoder features using pyramid camera prompts, and
(ii) \textit{a multi-query scale head} that estimates a global scene scale from depth cues and camera prompt.
Finally,
the DPT decoder is trained to output relative depth $\hat{\mathbf{d}}_{\mathrm{rel}}$ re-scaled to $[0,1]$,
and the scale head predicts a positive scale factor, \textit{i.e.}, $\mathbf{s}$, for metric calibration:
\begin{equation}
\hat{\mathbf{d}}_{\mathrm{metric}} = d_{\max}\,\big(\mathbf{s}\odot \hat{\mathbf{d}}_{\mathrm{rel}}\big).
\end{equation}
where $d_{\max}$ is a constant maximum depth and $\odot$ denotes multiplication.
We deliberately use a scale-only formulation in this work. This choice is not meant to imply that shift correction is universally unnecessary.
Rather, in the tiny regime, we find it preferable to keep metric calibration simple and stable,
so that adaptation does not interfere excessively with the transferred relative-depth prior.
We next detail the \textit{pyramid camera prompt}, \textit{camera adapter}, and \textit{multi-query scale head}.

\vspace{5pt}\noindent\textbf{Pyramid Camera Prompt.}
To make metric fine-tuning camera-aware without sacrificing efficiency,
we convert the camera intrinsics $\mathbf{K}\!\!=\!\!(f_x,f_y,c_x,c_y)$ into dense prompts for TinyViM.
Following the ray-direction prompt construction in prior work \cite{piccinelli2025unidepthv2},
we compute a per-pixel camera-ray field from $\mathbf{K}$ and apply a Fourier encoding to obtain a prompt.
Unlike single-resolution conditioning in fixed-resolution ViT pipelines,
TinyViM progressively downsamples features ($1/4\!\rightarrow\!1/8\!\rightarrow\!1/16\!\rightarrow\!1/32$),
where camera effects should be injected at matching resolutions.
Accordingly,
we build prompts $\{\mathbf{C}^{(s)}\!\!\in\!\!\mathbb{R}^{C_s\times H/s\times W/s}\}$ at strides $s\!\in\!\{4,8,16,32\}$,
and inject them at the corresponding stages.
In this way,
high-resolution prompts guide local structure adaptation in shallow layers,
while coarse prompts support global scale calibration in deep layers,
making camera-conditioned fine-tuning more stable for tiny models.

\vspace{5pt}\noindent\textbf{Camera Adapter.}
Given the pyramid prompts,
we inject them into each frozen encoder stage via a lightweight residual adapter.
Let $\mathbf{F}^{(l)}$ be the frozen feature map at stage $l$.
We compute an updated feature $\bar{\mathbf{F}}^{(l)}$ as
\begin{equation}
\bar{\mathbf{F}}^{(l)} = \mathbf{F}^{(l)} +  \mathrm{CA}\big(\mathbf{F}^{(l)},\,\mathrm{Proj}(\mathbf{C}^{(s)})\big),
\end{equation}
where $\mathrm{CA}$ is a 1-layer, 1-head cross-attention block
with queries from $\mathbf{F}^{(l)}$ and keys/values from $\mathrm{Proj}(\mathbf{C}^{(s)})$.
$\mathrm{Proj}(\cdot)$ applies a $3\times3$ depthwise convolution followed by a $1\times1$ projection,
without activation or normalization.
We keep this adapter intentionally small to preserve real-time inference while enabling camera-conditioned metric calibration.

\vspace{5pt}\noindent\textbf{Multi-Query Scale Head.}
Estimating a single global scale from a tiny backbone is usually unstable,
as the prediction may be affected by spurious textures or local outliers.
We therefore design a multi-query scale head that aggregates multiple global cues via an implicit ensemble.
Given the deepest feature $\bar{\mathbf{F}}^{(4)}$ and its aligned camera prompt $\mathbf{C}^{(32)}$,
we form a context set $\mathbf{X}=[\bar{\mathbf{F}}^{(4)},\mathbf{C}^{(32)}]$ by concatenation and apply a lightweight Transformer decoder with $Q$ learnable queries.
Specifically, with $\mathbf{Q}\in\mathbb{R}^{Q\times128}$ and a 2-layer, 4-head decoder $\mathrm{Dec}(\cdot)$, we compute the scale factor $\mathbf{s}$:
\begin{equation}
\mathbf{Z}=\mathrm{Dec}(\mathbf{Q},\mathbf{X}), \quad
\mathbf{s}=\frac{1}{Q}\sum\nolimits_{q=1}^{Q} \sigma \big(\mathrm{MLP}(\mathbf{Z}_{q,:})\big),
\end{equation}
where $\mathrm{MLP}(\cdot)$ is a two-layer MLP (128$\rightarrow$128$\rightarrow$1) with ReLU,
and $\sigma(\cdot)$ enforces a positive, bounded scale.
Different queries attend to complementary scale cues, \textit{e.g.}, scene layout/FoV-driven geometry/large planes.
The averaging reduces their variance,
yielding a more stable scale estimation.

\begin{table*}[t]
\centering
\setlength{\tabcolsep}{2pt}
\renewcommand{\arraystretch}{1.08}
\caption{Zero-shot affine-invariant depth estimation. Better: AbsRel$\downarrow$, $\delta_1\uparrow$. Best and second-best results are marked in bold and underlined.
DepthART is reported at three scales: S/B/L. All latency is measured on an RTX A6000 with batch size 1 under
strict FP32 execution with TF32 disabled,
and should thus be interpreted as deployment-oriented throughput rather than a strict same-resolution speed benchmark.
``*'' marks in-domain result. ``Params'' denotes the full model parameter amount.}
\resizebox{\linewidth}{!}{
\begin{tabular}{l |rrr|cc|cc|cc|cc|cc}
\toprule
\multirow{2}{*}{\textbf{Method}} & \multirow{2}{*}{\textbf{Encoder}} & \multirow{2}{*}{\textbf{Params}} & \textbf{Native Lat.} &
\multicolumn{2}{c|}{\textbf{NYUD v2}} &
\multicolumn{2}{c|}{\textbf{KITTI}} &
\multicolumn{2}{c|}{\textbf{ETH3D}} &
\multicolumn{2}{c|}{\textbf{DIODE-Full}} &
\multicolumn{2}{c}{\textbf{DDAD}} \\
 &  &  & \textbf{A6000} & 
\textbf{AbsRel}$\downarrow$ & $\mathbf{\delta_1} \uparrow$ &
\textbf{AbsRel}$\downarrow$ & $\mathbf{\delta_1} \uparrow$ &
\textbf{AbsRel}$\downarrow$ & $\mathbf{\delta_1} \uparrow$ &
\textbf{AbsRel}$\downarrow$ & $\mathbf{\delta_1} \uparrow$ &
\textbf{AbsRel}$\downarrow$ & $\mathbf{\delta_1} \uparrow$ \\
\midrule
\multirow{4}{*}{\makecell[l]{\textbf{MiDaS v3.1} \shortcite{ranftl2022midas}\\
\textit{\textbf{Comprehensive Models}}} } & LeViT & 51M & 3.42 ms  &
0.090* &	0.915* &	0.127* &	0.830* &	0.129 &	0.875 &	0.240 &	0.698 &	0.127 &	0.836   \\
& Swin2-T & 42M & 7.31 ms  &
0.075* &	0.942* &	0.103* &	0.883* &	0.119 &	0.886 &	0.229 &	0.718 &	0.117 &	0.848   \\
& Swin2-L & 213M & 51.47 ms  &
0.055* &	0.972* &	0.078* &	0.932* &	0.086 &	0.949 &	0.212 &	0.747 &	0.094 &	0.899  \\
 & Beit-L & 345M & 156.56 ms  &
0.047* &	0.980* &	0.130* &	0.843* &	0.072 &	\underline{0.961} &	\underline{0.198} &	0.762 &	0.092 &	0.908  \\

\midrule
\multirow{2}{*}{\makecell[l]{\textbf{Metric3D v2} \shortcite{hu2024metric3dv2}\\
\textbf{\textit{Large-scale Metric MDE}}}
}
& ViT-S & 37.5M & 74.99 ms  & 0.056 & 0.965 & \underline{0.064} & \underline{0.950} & \underline{0.062} & 0.955 & 0.247 & \underline{0.789} & -- & -- \\
& ViT-L & 411.9M & 452.54 ms  & \textbf{0.042} & \textbf{0.980} & \textbf{0.046} & \textbf{0.979} & \textbf{0.042} & \textbf{0.987} & \textbf{0.141} & \textbf{0.882} & \textbf{0.072} & \textbf{0.948} \\

\midrule
\multirow{3}{*}{\makecell[l]{\textbf{Depth Anything v2} \shortcite{depth_anything_v2}\\
\textbf{\textit{Large-scale Relative MDE}}}}
& ViT-S & 24.8M & 13.06 ms  & 0.053 & 0.973 & 0.078 & 0.936 & 0.142 & 0.851 & 0.228 & 0.751 & 0.085 & 0.921 \\
& ViT-B & 97.5M & 33.04 ms  & 0.049 & 0.976 & 0.078 & 0.939 & 0.137 & 0.858 & 0.236 & 0.756 &
\underline{0.081} & \underline{0.929} \\
& ViT-L & 335.3M & 102.46 ms  &
\underline{0.045} & \underline{0.979} & 0.074 & 0.946 & 0.131 & 0.865 & 0.230 & 0.758 & 0.089 & 0.926 \\

\midrule
\multirow{3}{*}{\makecell[l]{\textbf{AnyDepth} \shortcite{anydepth}\\
\textbf{\textit{Real-world Deployable}}}}
& ViT-S & 26.5M & --  & 0.082 & 0.932 & 0.102 & 0.883 & 0.084 & 0.935  & 0.247 & 0.714 & - & - \\
& ViT-B & 95.5M &  -- & 0.072 & 0.950 & 0.097 & 0.901 & 0.080 & 0.945 & 0.236 & 0.727 & - & - \\
& ViT-L & 313.4M &  -- & 0.060 & 0.968 & 0.086 & 0.926 & 0.096 & 0.954 & 0.226 & 0.736 & - & - \\

\midrule
\multirow{3}{*}{ \makecell[l]{\textbf{DepthART} (Ours)\\
\textbf{\textit{Foundation Tiny MDE}}
}}
& TinyViM-S & 6.0M  &  4.09 ms  & 0.059 & 0.964 & 0.082 & 0.930 & 0.084 & 0.950 & 0.214 & 0.745 & 0.095 & 0.900 \\
& TinyViM-B & 11.4M &  4.91 ms & 0.057 & 0.969 & 0.088 & 0.929 & 0.092 & 0.954 & 0.214 & 0.746 & 0.094 & 0.906 \\
& TinyViM-L & 32.6M & 8.07 ms  & 0.053 & 0.971 & 0.079 & 0.933 & 0.091 & 0.958 & 0.208 & 0.754 & 0.095 & 0.910 \\
\bottomrule
\end{tabular}}
\label{tab:zero_shot_relative_depth}
\end{table*}

\vspace{5pt}\noindent\textbf{Fine-tuning Phase.}
We perform CamFT on a single dataset (NYUD v2 or KITTI) with standard supervision.
In this process, we minimize the SiLog loss on valid pixels with a multi-scale edge loss \cite{ranftl2022midas}:
\begin{equation}
\mathcal{L}_{\mathrm{ft}}
=
\mathcal{L}_{\mathrm{SiLog}}(\hat{\mathbf{d}}_{\mathrm{metric}},\,\mathbf{d})
+\lambda\,\mathcal{L}_{\mathrm{Edge}}(\hat{\mathbf{d}}_{\mathrm{metric}},\,\mathbf{d}),
\end{equation}
where $\mathbf{d}$ is the metric ground truth and $\lambda$ weights the edge term.
During CamFT,
we freeze the \textit{distilled encoder} and update the \textit{camera adapters},
the \textit{multi-query scale head}, and \textit{DPT}, mitigating forgetting while calibrating metric scale.

\begin{table*}[t]
\centering
\setlength{\tabcolsep}{2pt}
\caption{
In-domain (\textcolor{pink!60}{pink}) and zero-shot (\textcolor{cyan!50}{cyan}) metric depth results without any re-scaling post-processing.
Models are fine-tuned on NYUD v2 or KITTI with metric supervision and directly tested on unseen datasets.
Best results are in \textbf{bold}. ``--'' denotes missing KITTI-fine-tuned checkpoints (e.g., FastDepth and DANet),
so DDAD/ETH3D/DIODE results are not reported.
Depth Anything v1-L is a heavy model.
}
\resizebox{\textwidth}{!}{
\begin{tabular}{l|
                >{\columncolor{pink!30}}c >{\columncolor{pink!30}}c
                >{\columncolor{cyan!15}}c >{\columncolor{cyan!15}}c
                >{\columncolor{cyan!15}}c >{\columncolor{cyan!15}}c
                >{\columncolor{cyan!15}}c >{\columncolor{cyan!15}}c|
                >{\columncolor{pink!30}}c >{\columncolor{pink!30}}c
                >{\columncolor{cyan!15}}c >{\columncolor{cyan!15}}c
                >{\columncolor{cyan!15}}c >{\columncolor{cyan!15}}c
                >{\columncolor{cyan!15}}c >{\columncolor{cyan!15}}c}
\toprule
\multirow{3}{*}{\textbf{Method}} &
\multicolumn{2}{>{\columncolor{pink!30}}c}{\textbf{\textit{In Domain}}} &
\multicolumn{6}{>{\columncolor{cyan!15}}c|}{\textbf{\textit{Zero-shot}}} &
\multicolumn{2}{>{\columncolor{pink!30}}c}{\textbf{\textit{In Domain}}} &
\multicolumn{6}{>{\columncolor{cyan!15}}c}{\textbf{\textit{Zero-shot}}}\\
&
\multicolumn{2}{>{\columncolor{pink!30}}c}{\textbf{NYUD v2}} &
\multicolumn{2}{>{\columncolor{cyan!15}}c}{\textbf{iBims-1}} &
\multicolumn{2}{>{\columncolor{cyan!15}}c}{\textbf{SUN RGB-D}} &
\multicolumn{2}{>{\columncolor{cyan!15}}c|}{\textbf{DIODE Indoor}} &
\multicolumn{2}{>{\columncolor{pink!30}}c}{\textbf{KITTI}} &
\multicolumn{2}{>{\columncolor{cyan!15}}c}{\textbf{DDAD}} &
\multicolumn{2}{>{\columncolor{cyan!15}}c}{\textbf{ETH3D Outdoor}} &
\multicolumn{2}{>{\columncolor{cyan!15}}c}{\textbf{DIODE Outdoor}} \\
& \textbf{$\delta_1 \uparrow$} & \textbf{RMSE $\downarrow$}
& \textbf{$\delta_1 \uparrow$} & \textbf{RMSE $\downarrow$}
& \textbf{$\delta_1 \uparrow$} & \textbf{RMSE $\downarrow$}
& \textbf{$\delta_1 \uparrow$} & \textbf{RMSE $\downarrow$}
& \textbf{$\delta_1 \uparrow$} & \textbf{RMSE $\downarrow$}
& \textbf{$\delta_1 \uparrow$} & \textbf{RMSE $\downarrow$}
& \textbf{$\delta_1 \uparrow$} & \textbf{RMSE $\downarrow$}
& \textbf{$\delta_1 \uparrow$} & \textbf{RMSE $\downarrow$} \\
\midrule
Depth Anything v1-L    & \textbf{0.984} & \textbf{0.206} & 0.639 & 0.965 & 0.647 & 0.585 & 0.291 & 2.504 & \textbf{0.982} & \textbf{1.90} & 0.787 & 8.214 & 0.164 & 6.507 & \textbf{0.288} & 8.994 \\
\midrule
FastDepth            & 0.771 & 0.604 & 0.429 & 1.607 & 0.560 & 0.736 & 0.275 & 2.778 & 0.825 & 4.33  & -- & --    & --    & --    & --    & --    \\
DANet                & 0.831 & 0.488 & 0.55 & 1.11 & 0.616 & 0.623 & 0.221 & 2.909 & 0.920 & 3.52  & -- & --    & --    & --    & --    & --    \\
GuideDepth       & 0.840 & 0.478 & 0.560 & 1.014 & 0.603 & 0.657 & 0.271 & 2.766 & 0.882 & 4.10  & 0.197 & 21.31  & 0.171 & 9.539 & 0.158 & 14.25 \\
\midrule
\textbf{DepthART-S} & 0.924 & 0.340 & 0.713 & 0.693 & 0.809 & 0.379 & 0.292 & \textbf{1.687} & 0.943 & 3.12  & 0.811 & 7.140 & 0.228 & 6.518 & 0.216 & 9.862 \\
\textbf{DepthART-B}
& 0.942 & 0.307 & 0.732 & 0.672
& 0.831 & 0.342 & 0.287 & 2.018
& 0.951 & 2.948 & 0.847 & 6.475
& 0.227 & 6.831 & 0.231 & 9.474 \\
\textbf{DepthART-L}
& 0.946 & 0.295 & \textbf{0.747} & \textbf{0.646}
& \textbf{0.847} & \textbf{0.326} & \textbf{0.298} & 2.012
& 0.956 & 2.784 & \textbf{0.867} & \textbf{6.340}
& \textbf{0.258} & \textbf{5.794} & 0.249 & \textbf{8.823} \\

\bottomrule
\end{tabular}}
\label{tab:camft_nyud_kitti_zeroshot}
\end{table*}

\section{Experiments}
\subsection{Dataset}
To maximize coverage in relative-depth pre-training, we aggregate data from 11 sources:
ApolloScapeExtra \cite{huang2018apolloscape},
BDD100K \cite{yu2020bdd100k},
Cityscapes \cite{cordts2016cityscapes},
DIML-indoor \cite{diml},
GoogleLandmark \cite{weyand2020googlelandmark},
Holopix50K \cite{hua2020holopix50k},
HRWSI \cite{hrwsi},
ImageNet21K \cite{imagenet},
LSUN \cite{yu15lsun},
Objects365 \cite{shao2019objects365},
and ReDWeb \cite{redweb}.
For the large-scale datasets (BDD100K, GoogleLandmark, ImageNet21K, LSUN, and Objects365),
we do not use the full collections but use our BRDS to extract bias-resistant samples, resulting in a \textit{1.7M} image training set.
After pre-training,
we evaluate zero-shot affine-invariant depth on NYUD v2 (Eigen crop) \cite{silberman2012NYUD}, KITTI (Eigen crop) \cite{kitti}, ETH3D \cite{schops2017eth3d}, DIODE \cite{diode}, and DDAD \cite{ddad}.
For metric-scale fine-tuning,
we train on a single metric dataset (NYUD v2 or KITTI with Eigen crop) and further report zero-shot metric transfer on iBims-1 \cite{ibims}, SUN RGB-D \cite{sunrgbd}, DIODE, DDAD, and ETH3D.

\subsection{Baselines}
For \textbf{\textit{affine-invariant}} depth, we compare with \textbf{\textit{MiDaS v3.1}}~\cite{ranftl2022midas}, which provides a broad family of generalizable relative MDE models from lightweight to heavyweight variants, including a fastest model MiDaS-LeViT; \textbf{\textit{Metric3D v2}}~\cite{hu2024metric3dv2}, as a representative generalizable metric MDE method; \textbf{\textit{Depth Anything v2}}~\cite{depth_anything_v2}, as both a representative generalizable relative MDE method and our teacher model; and \textbf{\textit{AnyDepth}}~\cite{anydepth}, which targets both lightweight deployment and cross-scene generalization. Since AnyDepth does not release full model code, we only report its parameter count and accuracy, not inference speed. We exclude diffusion-based methods because they are typically much slower, and self-supervised methods because they follow a different paradigm.

For \textbf{\textit{metric}} depth, we compare with the lightweight models \textbf{\textit{FastDepth}}~\cite{wofk2019fastdepth}, \textbf{\textit{DANet}}~\cite{sheng2022danet}, and \textbf{\textit{GuideDepth}}~\cite{guidedepth}. We do not include \textbf{\textit{TuMDE}}~\cite{tumde}, \textbf{\textit{LightDepth}}~\cite{lightdepth}, or \textbf{\textit{METER}}~\cite{meter}, because only partial code is publicly available and our reproduced results are substantially worse than those reported. We also include \textbf{\textit{Depth Anything v1-L}}~\cite{depthanything}, which, although not lightweight, is the only publicly available Depth Anything model aligned with our setting.
To keep the comparison fair, we only consider models fine-tuned on NYUD v2 or KITTI and tested on the remaining datasets.
Accordingly, we do not treat large-scale camera-aware metric MDE models such as UniDepth and Metric3D as direct baselines, since their training data, model capacity, and fine-tuning settings differ substantially from ours.
Instead, our metric evaluation measures whether a model can retain useful metric transfer under strict deployment-oriented constraints.
These large-scale models remain important related references,
but they are not strict ``apples-to-apples'' comparisons in our setting.

\begin{figure*}[t]
\centering
\includegraphics[width=\linewidth]{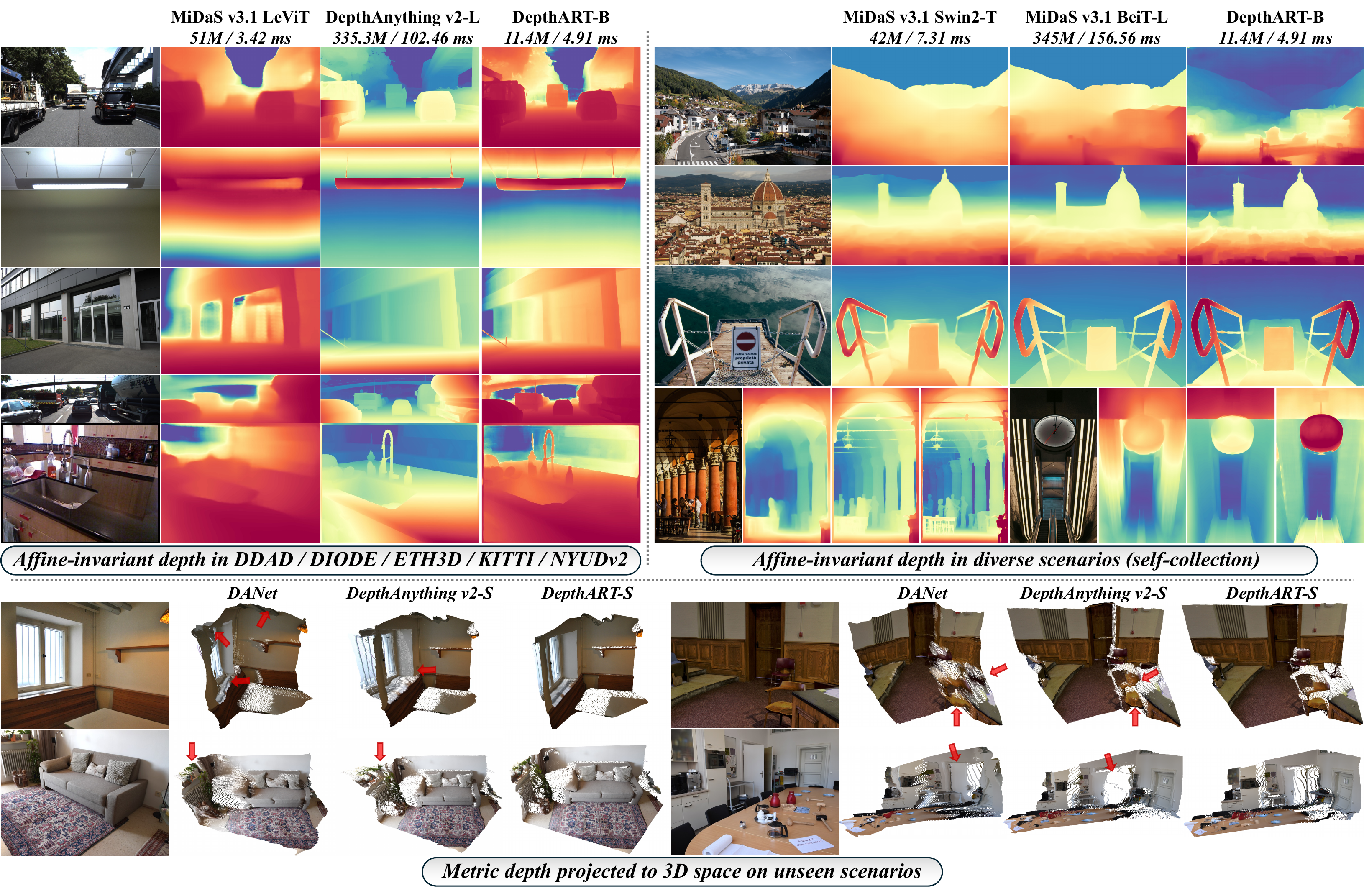}
\caption{Qualitative comparison of affine-invariant depth and metric 3D reconstruction.
Top-left: predictions on five zero-shot benchmarks
(DDAD, DIODE, ETH3D, KITTI, and NYUD v2).
Top-right: predictions on diverse self-collected scenes.
Different representative baselines are used across the two groups to cover both lightweight and large models under limited space.
DepthART-B more consistently preserves thin structures, object boundaries, and global scene layouts.
Bottom: metric predictions after NYUD v2 fine-tuning are projected directly into 3D without post-hoc scale or shift alignment and evaluated on unseen indoor scenes. DepthART-S produces more coherent geometry around windows, doors, chairs, and furniture boundaries than DANet and Depth Anything v2-s (fine-tuned on Hypersim).}
\label{fig:vis}
\end{figure*}

\subsection{Implementation Details}
DepthART is trained on an A800, while all desktop latency is measured on an RTX A6000 with batch size 1 under
strict FP32 execution with TF32 disabled.
The main accuracy comparisons evaluate DepthART at $448\times448$.
For the accuracy--efficiency analysis in Table~\ref{tab:efficiency_deployment},
we additionally report resolution-matched accuracy and efficiency at $224\times224$ and at each method's
standard operating resolution.
For BRDS, we use $J=8$
and $B=10$. Stage-1 distillation runs for 40 epochs with batch size 96 using Adam $(\beta_1,\beta_2)=(0.9,0.999)$, learning rate $1\mathrm{e}{-4}$, weight decay 0.01, and horizontal flips only.
We apply source-specific reduction factors $\gamma$ to the five large-scale datasets.
CamFT runs for 40 epochs with batch size 64 and initial learning rate $5\mathrm{e}{-5}$, with $Q=8$, $d_{\max}=10$ for NYUD v2, and $d_{\max}=80$ for KITTI.
We further apply mild intrinsics perturbation during CamFT by randomly rescaling images to $[0.9,1.1]$ and cropping $[0.8,1.0]$ of the resized area, which serves as regularization for small camera variations rather than severe intrinsics noise or missing metadata. RGB images use bilinear interpolation, while depth maps use nearest-neighbor interpolation.

\subsection{Quantitative and Qualitative Comparison}
\vspace{5pt}\noindent\textbf{Zero-Shot Affine-Invariant Depth Estimation.}
Table~\ref{tab:zero_shot_relative_depth} shows zero-shot affine-invariant results on five benchmarks spanning diverse scenes.
Across three lightweight scales (TinyViM-S/B/L),
DepthART delivers consistently strong generalization on the same level of latency,
outperforming early baselines (\textit{e.g.}, MiDaS) with orders-of-magnitude fewer parameters than ViT-L models.
Notably, TinyViM-L achieves the best ETH3D $\delta_1$ (0.958) among models below 50M parameters and improves AbsRel on DIODE-Full (0.208) over Depth Anything v2 ViT-L (0.230),
showing that our BRDS helps to yield highly generalizable relative-depth cues.
Although Metric3D v2 (ViT-L) attains the best accuracy,
DepthART provides a far lighter,
deployment-friendly alternative that better matches real-time, on-device constraints.

\vspace{5pt}\noindent\textbf{In-domain/Zero-shot Metric Depth Estimation.}
Table \ref{tab:camft_nyud_kitti_zeroshot} shows both in-domain results (fine-tuned on NYUD v2 or KITTI) and cross-dataset transfer without further adaptation.
Scaling DepthART from S to B/L generally improves both in-domain accuracy and cross-dataset transfer.
After NYUD v2 fine-tuning, DepthART-L achieves the best results on iBims-1 ($0.747/0.646$) and SUN RGB-D ($0.847/0.326$).
On DIODE-Indoor,
DepthART-L obtains the best $\delta_1$ ($0.298$), while DepthART-S achieves the lowest RMSE ($1.687$). After KITTI fine-tuning,
DepthART-L achieves the best results on DDAD ($0.867/6.340$) and ETH3D-Outdoor ($0.258/5.794$), and the lowest RMSE on DIODE-Outdoor ($8.823$).
Depth Anything v1-L retains the highest DIODE-Outdoor $\delta_1$ ($0.288$).
Overall, CamFT provides strong metric transfer across model scales while retaining competitive in-domain accuracy.


\vspace{5pt}\noindent\textbf{Qualitative results.}
Fig.~\ref{fig:vis} shows qualitative results for affine-invariant depth and metric 3D reconstruction. 
Across five zero-shot benchmarks and additional self-collected scenes,
DepthART-B yields more coherent depth layouts and cleaner local structures than prior lightweight models, while remaining visually competitive with much larger baselines, i.e., the teacher, Depth Anything v2-L. 
It better preserves thin structures, object boundaries, and relative depth in challenging regions, where other methods often suffer from over-smoothing or local distortions. 
After metric fine-tuning,
DepthART-S also produces more geometrically consistent 3D reconstructions on unseen indoor scenes, showing less distortion around the window and windowsill in the two left cases, and cleaner geometry for the wooden door and chair in the two right cases, compared with DANet and Depth Anything v2-S.

\subsection{Additional Analysis}

\begin{table}[t]
\centering
\setlength{\tabcolsep}{3.0pt}
\renewcommand{\arraystretch}{1.12}
\caption{
Ablation of subset selection under three Stage-1 distillation budgets (0.5M/1.0M/1.7M).
We compare BRDS with uniform random sampling and K-Means selection under the same distillation schedule.
Bold marks the best result within each budget (ties allowed).
Percentages show relative changes,
with \textcolor{red!70!black}{red} for degradation and \textcolor{green!50!black}{green} for improvement.
}
\resizebox{\linewidth}{!}{
\begin{tabular}{cl|*{2}{cc}}
\toprule
\multirow{2}{*}{\makecell{\textbf{Data}\\\textbf{Amount}}} &
\multirow{2}{*}{\makecell{\textbf{Sampling}\\\textbf{Strategy}}} &
\multicolumn{2}{c}{\textbf{NYUD v2}} &
\multicolumn{2}{c}{\textbf{KITTI}} \\
& & $\mathbf{\delta_1} \uparrow$ & \textbf{AbsRel}$\downarrow$
  & $\mathbf{\delta_1} \uparrow$ & \textbf{AbsRel}$\downarrow$ \\
\midrule

\multirow{3}{*}{\textbf{1.7M}}
& Random       & 0.957{\textcolor{red!70!black}{\scriptsize -0.7\%}} & 0.065{\textcolor{red!70!black}{\scriptsize +10.2\%}}
              & 0.905{\textcolor{red!70!black}{\scriptsize -2.7\%}} & 0.104{\textcolor{red!70!black}{\scriptsize +26.8\%}} \\
& K-Means       & 0.958{\textcolor{red!70!black}{\scriptsize -0.6\%}} & 0.063{\textcolor{red!70!black}{\scriptsize +6.8\%}}
              & 0.923{\textcolor{red!70!black}{\scriptsize -0.8\%}} & 0.086{\textcolor{red!70!black}{\scriptsize +4.9\%}} \\
& \textbf{Bias-Resistant}  & \textbf{0.964} & \textbf{0.059}
              & \textbf{0.930} & \textbf{0.082} \\
\midrule

\multirow{3}{*}{\textbf{1.0M}}
& Random       & 0.947{\textcolor{red!70!black}{\scriptsize -1.4\%}} & 0.072{\textcolor{red!70!black}{\scriptsize +14.3\%}}
              & 0.903{\textcolor{red!70!black}{\scriptsize -2.3\%}} & 0.107{\textcolor{red!70!black}{\scriptsize +20.2\%}} \\
& K-Means       & 0.958{\textcolor{red!70!black}{\scriptsize -0.2\%}} & \textbf{0.063}{\textcolor{gray!70}{\scriptsize +0.0\%}}
              & 0.917{\textcolor{red!70!black}{\scriptsize -0.8\%}} & 0.090{\textcolor{red!70!black}{\scriptsize +1.1\%}} \\
& \textbf{Bias-Resistant}  & \textbf{0.960} & \textbf{0.063}
              & \textbf{0.924} & \textbf{0.089} \\
\midrule

\multirow{3}{*}{\textbf{0.5M}}
& Random       & 0.941{\textcolor{red!70!black}{\scriptsize -1.3\%}} & 0.076{\textcolor{red!70!black}{\scriptsize +11.8\%}}
              & 0.891{\textcolor{red!70!black}{\scriptsize -3.0\%}} & 0.117{\textcolor{red!70!black}{\scriptsize +30.0\%}} \\
& K-Means       & \textbf{0.954}{\textcolor{green!50!black}{\scriptsize +0.1\%}} & \textbf{0.067}{\textcolor{green!50!black}{\scriptsize -1.5\%}}
              & 0.911{\textcolor{red!70!black}{\scriptsize -0.9\%}} & 0.095{\textcolor{red!70!black}{\scriptsize +5.6\%}} \\
& \textbf{Bias-Resistant}  & 0.953 & 0.068
              & \textbf{0.919} & \textbf{0.090} \\
\bottomrule
\end{tabular}}
\label{tab:ablation_sampling}
\end{table}

\vspace{5pt}\noindent\textbf{Ablations for Sampling Strategy.}
We ablate BRDS by distilling our model with the same sampling amount using Random, K-Means, or BRDS,
under three budgets (0.5M/1.0M/1.7M).
Table \ref{tab:ablation_sampling} reports zero-shot affine-invariant depth results on NYUD v2 and KITTI.
At 1.7M it achieves the best accuracy on both datasets (NYUD v2 $0.964/0.059$, KITTI $0.930/0.082$),
and under tighter budgets it remains more sample-efficient,
notably improving KITTI over Random at 0.5M ($\delta_1$: $0.891\!\rightarrow\!0.919$, AbsRel: $0.117\!\rightarrow\!0.090$). These gains indicate that BRDS reduces dominant-mode overfitting and improves generalization with limited data.
At 0.5M, BRDS is slightly behind K-Means on some metrics,
as the budget is too small for the student to fully fit the more diverse, long-tail exposure emphasized by BRDS.
With sufficient data (1.0M+), BRDS becomes consistently superior.

\vspace{5pt}\noindent\textbf{CamFT Design Ablations.}
Table~\ref{tab:ablate_finetune_stage} isolates three factors:
freezing the distilled encoder (\textbf{Frz}),
the camera adapter (\textbf{CamAdp}),
and the Multi-Query Scale Head (\textbf{MQSH}).
Without freezing,
the model attains high in-domain accuracy on NYUD v2 (full fine-tuning: $0.928/0.328$, or even $0.930/0.333$ with CamAdp+MQSH) but transfers poorly to iBims-1 ($0.572/0.844$ and $0.613/0.779$),
showing severe cross-domain degradation.
With the trunk frozen,
removing either module causes a large drop on iBims-1 (w/o CamAdp: $0.534/0.879$; w/o MQSH: $0.544/0.866$),
while using both achieves the best transfer ($0.713/0.693$) with comparable NYUD v2 results ($0.924/0.340$).

\begin{table}[t]
\centering
\small
\setlength{\tabcolsep}{4.5pt}
\caption{CamFT Design ablations.
Models are fine-tuned on NYUD v2 and evaluated in-domain (ID) on NYUD v2 and zero-shot (ZS) on iBims-1.}
\begin{tabular}{ccc cc cc}
\toprule
\multirow{2}{*}{\textbf{Frz}}
& \multirow{2}{*}{\textbf{CamAdp}}
& \multirow{2}{*}{\textbf{MQSH}}
& \multicolumn{2}{c}{\textbf{NYUD v2 (ID)}}
& \multicolumn{2}{c}{\textbf{iBims-1 (ZS)}} \\
& & & \textbf{$\delta_1 \uparrow$} & \textbf{RMSE $\downarrow$}
      & \textbf{$\delta_1 \uparrow$} & \textbf{RMSE $\downarrow$} \\
\midrule
\xmark & \xmark & \xmark & 0.928 & 0.328 & 0.572 & 0.844 \\
\cmark & \xmark & \xmark & 0.887 & 0.388 & 0.562 & 0.877 \\
\cmark & \cmark & \xmark & 0.896 & 0.380 & 0.544 & 0.866 \\
\cmark & \xmark & \cmark & 0.908 & 0.365 & 0.534 & 0.879 \\
\rowcolor{gray!20}
\cmark & \cmark & \cmark & 0.924 & 0.340 & \textbf{0.713} & \textbf{0.693} \\
\xmark & \cmark & \cmark & \textbf{0.930} & 0.333 & 0.613 & 0.779 \\
\bottomrule
\end{tabular}
\label{tab:ablate_finetune_stage}
\end{table}

\begin{table}[t]
\centering
\small
\setlength{\tabcolsep}{5pt}
\renewcommand{\arraystretch}{1.15}
\caption{Ablations for adapter design.
We study (i) how to fuse image features with camera prompts and (ii) where to insert the adapter.
Cross-attention in the encoder yields the best in-domain accuracy on NYUD v2 and the strongest zero-shot transfer to iBims-1.}
\resizebox{\linewidth}{!}{
\begin{tabular}{lcccc}
\toprule
\multirow{2}{*}{\textbf{Adapter Design Choices}} &
\multicolumn{2}{c}{\textbf{NYUD v2 (ID)}} &
\multicolumn{2}{c}{\textbf{iBims-1 (ZS)}} \\
& $\delta_1 \uparrow$ & \textbf{RMSE} $\downarrow$ & $\delta_1 \uparrow$ & \textbf{RMSE} $\downarrow$ \\
\midrule
\multicolumn{5}{l}{\textbf{\textit{Fusion method of camera prompt and feature}}} \\
Additive
& 0.911{\textcolor{red!70!black}{\scriptsize -1.4\%}} & 0.361{\textcolor{red!70!black}{\scriptsize +6.2\%}}
& 0.568{\textcolor{red!70!black}{\scriptsize -20.3\%}} & 0.840{\textcolor{red!70!black}{\scriptsize +21.2\%}} \\
Concat + Proj
& 0.913{\textcolor{red!70!black}{\scriptsize -1.2\%}} & 0.351{\textcolor{red!70!black}{\scriptsize +3.2\%}}
& 0.616{\textcolor{red!70!black}{\scriptsize -13.6\%}} & 0.781{\textcolor{red!70!black}{\scriptsize +12.7\%}} \\
\textbf{Cross-Attn}
& \textbf{0.924} & \textbf{0.340}
& \textbf{0.713} & \textbf{0.693} \\
\midrule
\multicolumn{5}{l}{\textbf{\textit{Adapter insertion position}}} \\
Decoder-side adapter
& 0.914{\textcolor{red!70!black}{\scriptsize -1.1\%}} & 0.346{\textcolor{red!70!black}{\scriptsize +1.8\%}}
& 0.611{\textcolor{red!70!black}{\scriptsize -14.3\%}} & 0.827{\textcolor{red!70!black}{\scriptsize +19.3\%}} \\
\textbf{Encoder-side adapter}
& \textbf{0.924} & \textbf{0.340}
& \textbf{0.713} & \textbf{0.693} \\
\bottomrule
\end{tabular}
}
\label{tab:ablation_adapter_design}
\end{table}

\begin{table*}[t]
\centering
\caption{
Accuracy--efficiency comparison on an RTX A6000 and on-device
deployment results on a Jetson Orin NX.
(a) reports efficiency under a controlled $224^2$ setting and under
the default operating settings of each method.
(b) reports FP32 relative-depth inference on a Jetson Orin NX 8GB
under MAXN mode.
The best and second-best values are highlighted in bold and underlined,
respectively.
}
\label{tab:efficiency_deployment}

\noindent
\begin{minipage}[t]{0.62\textwidth}
\vspace{0pt}
\centering

\textbf{(a) RTX A6000 48GB}
\quad
\textit{Strict FP32}

\vspace{0.6mm}

{
\setlength{\tabcolsep}{1.0pt}
\renewcommand{\arraystretch}{0.95}
\setlength{\aboverulesep}{0.25ex}
\setlength{\belowrulesep}{0.25ex}
\setlength{\cmidrulesep}{0.15ex}

\resizebox{\linewidth}{!}{%
\begin{tabular}{
l c | *{5}{c} | *{5}{c}
}
\toprule

\multirow{3.5}{*}{\textbf{Method}}
&
\multirow{3.5}{*}{\headtwo{Params}{M}}
&
\multicolumn{5}{c|}{
  \grouptwo{Controlled Setting}{All inputs are $224^2$}
}
&
\multicolumn{5}{c}{
  \grouptwo{Native Setting}{Swin2-T: $256^2$; DA-v2: $518^2$; DepthART: $448^2$}
}
\\

\cmidrule(lr){3-7}
\cmidrule(lr){8-12}

&
&
\headtwo{NYUD}{$\delta_1\uparrow$}
&
\headtwo{FLOPs}{G$\downarrow$}
&
\headtwo{Mem.}{MB$\downarrow$}
&
\headtwo{M-Lat.}{ms$\downarrow$}
&
\headtwo{E2E}{ms$\downarrow$}
&
\headtwo{NYUD}{$\delta_1\uparrow$}
&
\headtwo{FLOPs}{G$\downarrow$}
&
\headtwo{Mem.}{MB$\downarrow$}
&
\headtwo{M-Lat.}{ms$\downarrow$}
&
\headtwo{E2E}{ms$\downarrow$}
\\

\midrule

MiDaS v3.1-LeViT
& 51.0
& 0.915\textsuperscript{*}
& 5.3755
& 213.31
& 3.42
& 10.10
& \multicolumn{5}{>{\columncolor{statusgray}}c}{
  \textcolor{black}{\textbf{N/A}}
}
\\

MiDaS v3.1-Swin2-T
& 42.0
& \multicolumn{5}{>{\columncolor{statusgray}}c|}{
  \textcolor{black}{\textbf{N/A}}
}
& 0.942\textsuperscript{*}
& 27.58
& 242.65
& 7.31
& 13.66
\\

\midrule

Depth Anything v2-S
& 24.8
& \NAcell
& 7.735
& 125.82
& 4.83
& 10.05
& 0.973
& 41.304
& 228.60
& 13.06
& 21.70
\\

Depth Anything v2-B
& 97.5
& \NAcell
& 29.874
& 515.85
& 8.52
& 14.08
& \second{0.976}
& 159.507
& 694.55
& 33.04
& 40.77
\\

Depth Anything v2-L
& 335.3
& \NAcell
& 109.778
& 1457.19
& 20.89
& 25.88
& \best{0.979}
& 585.986
& 1770.14
& 102.46
& 145.71
\\

\midrule

DepthART-S
& 6.0
& 0.953
& \best{1.749}
& \best{46.36}
& \best{2.88}
& \best{3.72}
& 0.964
& \best{6.996}
& \best{91.15}
& \best{4.09}
& \best{6.98}
\\

DepthART-B
& 11.4
& \second{0.959}
& \second{2.348}
& \second{67.21}
& \second{3.36}
& \second{4.19}
& 0.969
& \second{9.394}
& \second{113.63}
& \second{4.91}
& \second{7.75}
\\

DepthART-L
& 32.6
& \best{0.967}
& 6.178
& 152.29
& 5.23
& 6.08
& 0.971
& 24.711
& 210.39
& 8.07
& 10.93
\\

\bottomrule
\end{tabular}%
}
}

\end{minipage}%
\hspace{0.005\textwidth}%
\begin{minipage}[t]{0.37\textwidth}
\vspace{0pt}
\centering

\textbf{(b) Orin NX 8GB}
\quad
\textit{FP32 with TF32 enabled, MAXN}

\vspace{0.6mm}

{
\setlength{\tabcolsep}{0.9pt}
\renewcommand{\arraystretch}{1.08}

\resizebox{\linewidth}{!}{%
\begin{tabular}{
l
c
c
c
c
}
\toprule

\textbf{Model}
&
\textbf{Input Res.}
&
\headtwo{Peak Mem.}{MB$\downarrow$}
&
\headtwo{M-Lat.}{ms$\downarrow$}
&
\headtwo{E2E}{ms$\downarrow$}
\\

\midrule


Depth Anything v2-S
&
$512^2$
&
124.71
&
126.40
&
127.15
\\

DepthART-S
&
$448^2$
&
\textbf{59.29}
&
\textbf{29.48}
&
\textbf{40.40}
\\

DepthART-B
&
$448^2$
&
\underline{79.84}
&
\underline{35.35}
&
\underline{46.09}
\\

DepthART-L
&
$448^2$
&
162.74
&
61.22
&
67.10
\\

\midrule

\addlinespace[2pt]

Depth Anything v2-S
&
$228^2$
&
113.38
&
18.94
&
21.65
\\
DepthART-S & $224^2$ & \textbf{50.68} & \textbf{9.80} & \textbf{12.43}
\\
DepthART-B & $224^2$ & \underline{71.23} & \underline{11.59} & \underline{14.20}
\\
DepthART-L & $224^2$ & 152.72 & 18.02 & 20.63
\\

\bottomrule
\end{tabular}
}
}

\end{minipage}

\vspace{1mm}

\begin{minipage}{0.99\textwidth}
\footnotesize
\textit{Notes.}
M-Lat.\ and E2E denote model-only and end-to-end latency, respectively.
Mem.\ denotes peak CUDA-allocated memory.
N/A indicates that the corresponding result is unavailable.
OOM indicates that inference was terminated because of insufficient memory.
\textsuperscript{*} indicates that the model was trained or fine-tuned on the corresponding target dataset, rather than evaluated in a zero-shot manner.
FPS is omitted because it is directly derived as $1000/\text{M-Lat.}$.
\end{minipage}

\end{table*}

\begin{figure*}[t]
    \centering
    \includegraphics[width=\textwidth]{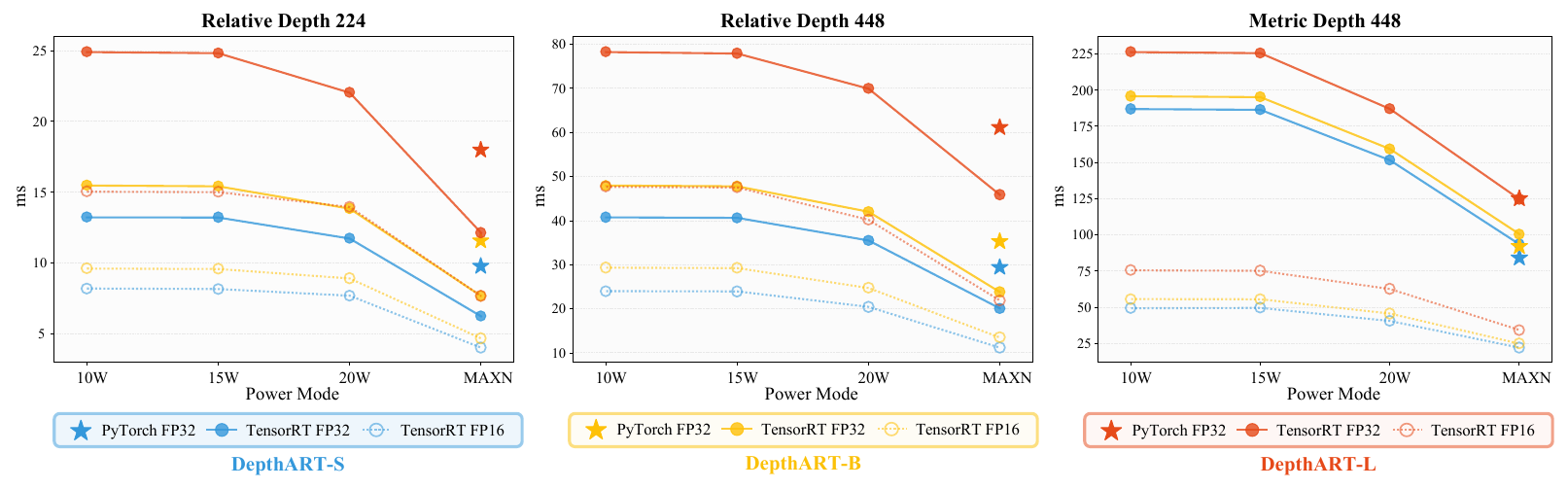}
    \caption{
    Model-only inference latency of DepthART on a Jetson Orin NX 8GB under different power modes.
    The three panels report relative-depth inference at $224^2$ and $448^2$, and metric-depth inference at $448^2$, respectively.
    Each color represents one model scale (S, B, or L).
    TensorRT FP32 and TensorRT FP16 results are shown across the 10W, 15W, 20W, and MAXN modes, while PyTorch FP32 results are reported at MAXN only.
    FP32 execution enables TF32 acceleration.
    Lower latency is better.
    }
    \label{fig:orin_nx_power}
\end{figure*}

\vspace{5pt}\noindent\textbf{Adapter Design Choices.}
Table \ref{tab:ablation_adapter_design} evaluates the camera adapter design along two axes:
fusion operator and insertion location.
Replacing cross-attention with simple fusion between camera prompts and image features leads to clear degradation, especially under domain shift:
on iBims-1, \textbf{Additive} drops from $0.713/0.693$ to $0.568/0.840$ and \textbf{Concat+Proj} to $0.616/0.781$ ($\delta_1$/RMSE);
even on NYUD v2, performance decreases to $0.911/0.361$ (Additive) and $0.913/0.351$ compared to $0.924/0.340$.
For insertion, placing the adapter on the decoder also hurts transfer, reducing iBims-1 from $0.713/0.693$ to $0.611/0.827$ (NYUD v2: $0.924/0.340 \rightarrow 0.914/0.346$).
Therefore, we adopt an encoder-side cross-attention adapter in all experiments.


\vspace{5pt}\noindent\textbf{Efficiency Analysis.}
Table~\ref{tab:efficiency_deployment} reports strict-FP32 efficiency on an RTX A6000 and on-device deployment on a Orin NX 8GB.
At $224^2$, DepthART-S/B/L achieve 347/298/191 FPS with model-only latencies of 2.88/3.36/5.23\,ms and end-to-end latencies of 3.72--6.08\,ms.
At $448^2$, they retain 245/204/124 FPS with 4.09--8.07\,ms model-only and 6.98--10.93\,ms end-to-end latency.
At their standard resolutions,
DepthART-S/B/L are respectively 3.2$\times$/6.7$\times$/12.7$\times$ faster than the corresponding Depth Anything V2 variants in model-only inference,
while requiring substantially fewer FLOPs and less memory.


On the Orin NX, the relative-depth S/B/L models reach
$102/86/55$ FPS at $224^2$ and $34/28/16$ FPS at $448^2$ in FP32 with TF32 enabled,
with end-to-end latencies of $12.43$--$20.63$\,ms and
$40.40$--$67.10$\,ms, respectively.
Figure~\ref{fig:orin_nx_power} further profiles relative- and metric-depth inference under the 10W, 15W, 20W, and MAXN modes.
TensorRT consistently reduces latency across model scales and resolutions, with FP16 providing the highest throughput, while higher power modes yield further acceleration.
Although incomplete Jetson Nano 4GB measurements are omitted from the table, DepthART-S still exceeds 15 FPS for $224^2$ relative-depth inference in strict FP32 on it.


\section{Conclusion}
We present \textbf{DepthART}, a tiny monocular depth model that transfers much of the generalization ability of foundation MDE models to deployment-oriented backbones, while supporting stable metric calibration through encoder-frozen camera-conditioned fine-tuning.
By combining bias-resistant sampling with camera-conditioned fine-tuning,
DepthART improves zero-shot generalization and metric accuracy over prior lightweight baselines while maintaining high efficiency across a scalable model family.

\vspace{5pt}\noindent\textbf{Limitations and future work.}
(1) CamFT still follows the Depth Anything fine-tuning setting.
Large-scale metric-depth training, \textit{e.g.}, in the spirit of UniDepth/Metric3D, is a key step to further strengthen metric-scale reliability. 
(2) CamFT assumes known camera intrinsics.
We will explore intrinsics estimation, \textit{e.g.}, GeoCalib \cite{veicht2024geocalib}, to reduce this reliance and enable metric depth without metadata.


\bibliographystyle{ACM-Reference-Format}
\bibliography{main}










\end{document}